\documentclass{article}

\usepackage{hyperref}

\usepackage{caption}
\usepackage{microtype}
\usepackage{graphicx}
\usepackage{booktabs} 
\usepackage{multirow}
\usepackage{slashbox}
\usepackage{mathtools}
\usepackage{listings}
\usepackage{xcolor}
\usepackage[draft]{minted}

\usepackage{tcolorbox}
\tcbuselibrary{minted, breakable,xparse,skins}

\definecolor{bg}{gray}{1}
\DeclareTCBListing{mintedbox}{O{}m!O{}}{%
  breakable=false,
  listing engine=minted,
  listing only,
  minted language=#2,
  minted style=default,
  minted options={%
    linenos,
    gobble=0,
    breaklines=true,
    breakafter=,,
    fontsize=\small,
    numbersep=15pt,
    #1},
  boxsep=0pt,
  left skip=0pt,
  right skip=0pt,
  left=25pt,
  right=0pt,
  top=3pt,
  bottom=3pt,
  arc=5pt,
  leftrule=0pt,
  rightrule=0pt,
  bottomrule=2pt,
  toprule=2pt,
  colback=bg,
  colframe=gray!70,
  enhanced,
  overlay={%
    \begin{tcbclipinterior}
    \fill[gray!20!white] (frame.south west) rectangle ([xshift=13pt]frame.north west);
    \end{tcbclipinterior}},
  #3}

\usepackage[accepted]{sysml2019}
\usepackage{amssymb}

\sysmltitlerunning{HyFed: A Hybrid Federated Framework for Privacy-preserving Machine Learning}

\begin{document}

\twocolumn[
\sysmltitle{HyFed: A Hybrid Federated Framework for Privacy-preserving Machine Learning}



\sysmlsetsymbol{equal}{*}

\begin{sysmlauthorlist}
\sysmlauthor{Reza Nasirigerdeh}{tum1,tum2}
\sysmlauthor{Reihaneh Torkzadehmahani}{tum1}
\sysmlauthor{Julian Matschinske}{uhh}
\sysmlauthor{Jan Baumbach}{uhh,sdu}
\sysmlauthor{Daniel Rueckert}{tum1,tum2,imperial,equal}
\sysmlauthor{Georgios Kaissis}{tum1,imperial,openmined,equal}

\end{sysmlauthorlist}
\sysmlaffiliation{tum1}{AI in Medicine and Healthcare, Technical University of Munich, Munich, Germany}
\sysmlaffiliation{tum2}{Klinikum rechts der Isar, Technical University of Munich, Munich, Germany}
\sysmlaffiliation{uhh}{Chair of Computational Systems Biology, University of Hamburg, Hamburg, Germany}
\sysmlaffiliation{sdu}{Department of Mathematics and Computer Science, University of Southern Denmark, Odense, Denmark}
\sysmlaffiliation{openmined}{OpenMined, Oxford, UK}
\sysmlaffiliation{imperial}{Biomedical Image Analysis Group, Imperial College London, London, UK}


\sysmlkeywords{HyFed, Machine Learning, Federated Learning}

\vskip 0.3in

\begin{abstract}
Federated learning (FL) enables multiple clients to jointly train a global model under the coordination of a central server. Although FL is a privacy-aware paradigm, where raw data sharing is not required, recent studies have shown that FL might leak the private data of a client through the model parameters shared with the server or the other clients. In this paper, we present the \textit{HyFed} framework, which enhances the privacy of FL while preserving the utility of the global model. \textit{HyFed} provides developers with a generic API to develop federated, privacy-enhanced algorithms. \textit{HyFed} supports both simulation and federated operation modes and its source code is publicly available at https://github.com/tum-aimed/hyfed.
\end{abstract}
]



\printAffiliationsAndNotice{\sysmlEqualContribution} 

\section{Introduction}
\label{sec:introduction}
\textit{Federated learning} (FL)~\cite{mcmahan2017communication} is a distributed learning approach, where multiple clients collaboratively train a global model under the orchestration of a central server without moving their private data off-site. In FL, the clients calculate the local model (e.g. gradients) from the data and global model and share only the local model with the server for aggregation~\cite{kairouz2019federated3}. Although FL provides a \textit{privacy-aware} training paradigm in which raw data sharing is not required, recent studies have demonstrated that FL might lead to privacy leakage in which the sensitive data of a client might be reconstructed using the parameters of the local models shared with the server or the other clients~\cite{fredrikson2015leakage1, melis2019leakage2, zhu2020leakage3,nasirigerdeh2021federatedpipelines}.

To enhance the privacy of FL, a \textit{privacy-enhancing technology} (PET) such as \textit{differential privacy} (DP)~\cite{dwork2006differentialprivacy}, \textit{secure multi-party computation} (SMPC)~\cite{cramer2015smpc}, or \textit{homomorphic encryption} (HE)~\cite{gentry2009homomorphicencryption} can be employed to hide the original values of the local model parameters from the server. We refer to a FL approach combined with a PET as a \textit{hybrid federated} approach in general and a FL-DP, FL-SMPC, or FL-HE approach in particular. Improving the privacy of FL comes with a cost of adversely affecting the performance/utility (FL-DP), network communication efficiency (FL-SMPC), or computational efficiency (FL-HE) of the underlying FL approach~\cite{torkzadehmahani2020survey}.

There are several open-source frameworks for federated learning including \textit{FedML}~\cite{chaoyanghe2020fedml}, \textit{Flower}~\cite{beutel2020flower}, \textit{TensorFlow Federated}~\cite{TFF}, \textit{FATE}~\cite{Fate} , \textit{PySyft}~\cite{PySyft}, and \textit{Sherpa.ai}~\cite{rodriguez2020sherpaai}. These frameworks have their own limitations: \textit{FedML} and \textit{Flower} do not offer any PET by default, and consequently, they do not address the privacy concerns of FL. The \textit{TensorFlow Federated} and \textit{Sherpa.ai} frameworks support the simulation mode but not truly federated mode. The core API (\textit{application programming interface}) offered by the \textit{FATE} framework is mainly driven by the custom scripts, implying that the developers must heavily modify the original source code to develop their own algorithms~\cite{kholod2021federatediot}. The focus of \textit{PySyft} is on deep learning (DL) and remote data science and its API currently allows for developing DL-based algorithms. 

\begin{figure*}[!ht]
    \centering
    \begin{minipage}{.8\textwidth}
        \centering
        \includegraphics[width=1\textwidth]{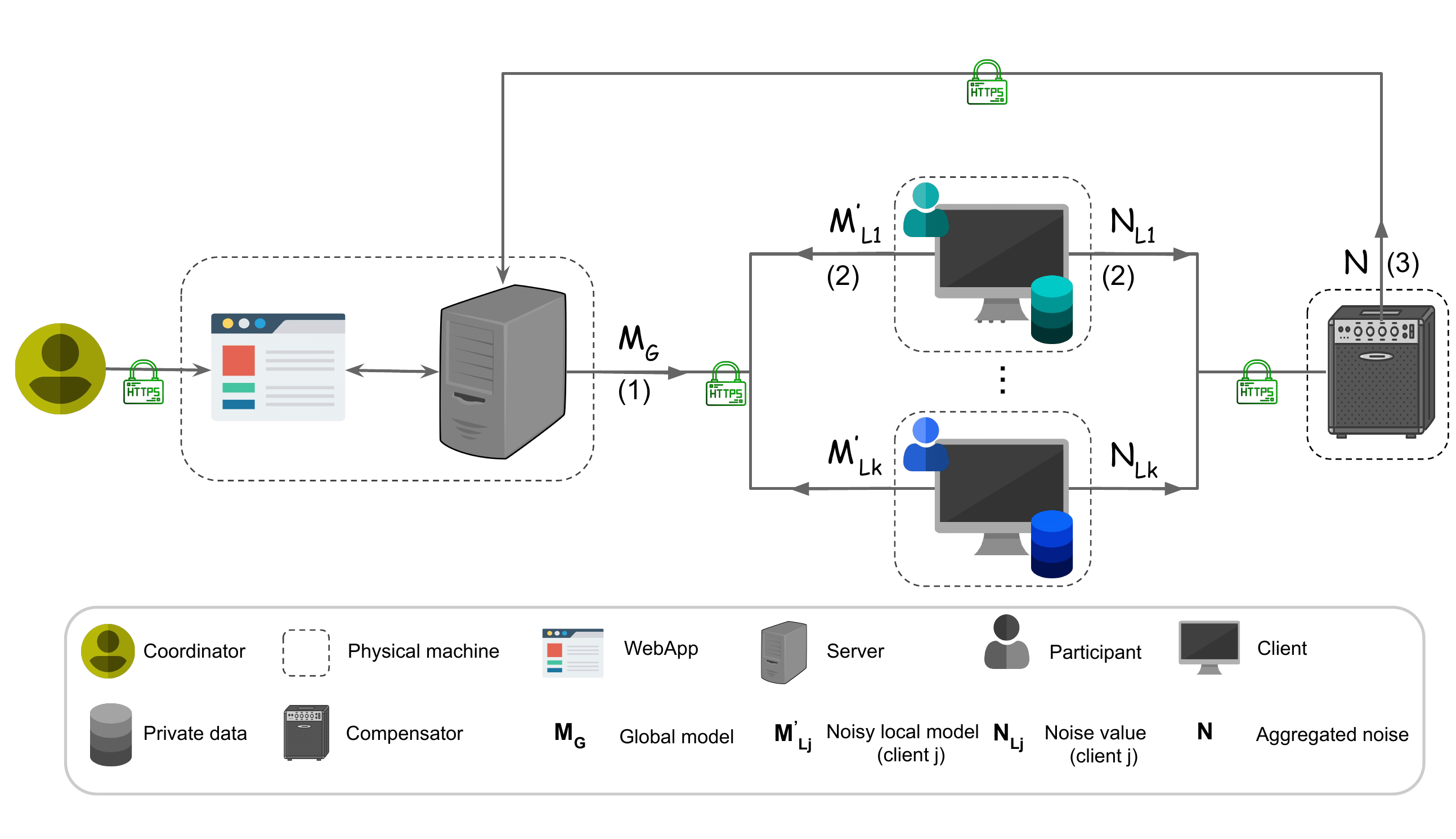}
    \end{minipage}
     \caption{\textit{HyFed} architecture: WebApp, client, compensator, and server are the main components of \textit{HyFed}. In the federated mode, the WebApp and server are running in the same physical machine while the clients and compensator are installed on separate machines. The communication between the machines is performed over a secure channel using the \textit{HTTPS} protocol. In each communication round: (1) the clients obtain the global model from the server, (2) compute the local models, mask the local models with noise, and share the noisy models and noise values with the server and compensator, respectively, (3) the compensator aggregates the noise values and sends the aggregated noise to the server, which in turn, updates the global model using the noisy local models and the negative of the aggregated noise.
     }
    \label{fig:hyfed-arch}
\end{figure*}

In this paper, we present a novel open-source framework called \textbf{\textit{HyFed}} (\textbf{Hy}brid \textbf{Fed}erated) for developing federated, privacy-enhanced machine learning algorithms. \textit{HyFed} consists of four software components: (1) \textbf{\textit{WebApp}} to set the hyper-parameters (e.g. learning rate) of the federated algorithm, (2) \textbf{\textit{client}} to compute the local model and mask it with noise, (3) \textbf{\textit{compensator}} to aggregate the noise values from the clients and share the aggregated noise with the server, and (4) \textbf{\textit{server}} to coordinate the training process and calculate the global model by adding up the noisy local models from the clients and the negative of the aggregated noise from the compensator. \textit{HyFed} enhances the privacy of FL without adversely impacting the performance of the global model because the negative of the aggregated noise cancels out the noise values added by the clients.  \textit{HyFed} is a generic framework, which enables developers to implement their own federated algorithms (e.g. regression, chi-square, deep learning, and etc) using its generic API.  \textit{HyFed} supports both simulation and federated modes of operation and its source code is publicly available at \textit{https://github.com/tum-aimed/hyfed}.

\section{The HyFed framework}
\label{sec:hyfed}
In this section, we elaborate on the architecture, functional workflow, masking mechanism, \textit{API}, and privacy of the \textit{HyFed} framework. Before delving into the details, we describe the terminology leveraged throughout the paper. \textit{User} is a human entity who interacts with the system. Each user is associated with a user \textit{account} specified by a unique username and password. \textit{Project} is a basic data entity bundling a unique project \textit{ID}, project name and description, federated tool name, algorithm, and the hyper-parameters of the algorithm. A user can have a \textit{coordinator} or \textit{participant} role or both. The coordinator creates the project and sets the hyper-parameters of the federated algorithm, and the participant contributes the private dataset to local model computation. \textit{Token} is a unique, random string assigned to each participant of the project for authentication purposes. The \textit{global model} is a set of parameters calculated by the server and shared with the clients whereas the \textit{local model} is a set of parameters computed by the clients from the local datasets and the global model. \textit{Noise} is a set of pseudo-random values with the same form as the local model. The \textit{noisy local model} is the sum of the local model and noise, which is calculated using \textit{modular arithmetic} for non-negative integers and ordinary addition for negative integers and floating-point numbers (see subsection \ref{subsec:masking-mech}).

\subsection{Architecture}
\label{subsec:hyfed-architecture}
The \textit{HyFed} architecture (Figure \ref{fig:hyfed-arch}) contains four components: WebApp, client, compensator, and server. The WebApp component provides users with a Web GUI (graphical user interface) to sign up an account and login/logout to/from the account. It also enables the coordinator to create the project, initialize the algorithm hyper-parameters, and create a token for each participant of the project. The users can see the list of projects as well as the summary, progress, and final results of the project through WebApp too. 

The client component (I) provides the participants with a simple GUI to join the project, view project information (e.g. name, description, algorithm hype-parameters, and etc), select the dataset file(s) to contribute to the project, and keep track the progress of the project, (II) computes the local model from the private data and global model, (III) mask the local model with noise to obtain the noisy local model, and (IV) shares the noisy local model with server and the noise with compensator.

The compensator is a lightweight component responsible for preserving the utility of the global model while the clients make the local models noisy to hide the real values of the models from the server. To this end, the compensator (I) receives the noise values from each client, (II) computes the aggregated noise by taking the sum over the noise values of the clients, and (III) sends the aggregated noise to server. 

The server component is the heart of the framework, which (I) implements the back-end functions to handle the WebApp's requests for creating the project/token or viewing the project summary/progress/results; (II) stores the user account information and project parameters in the database; (III) authenticates the users and the other components; (IV) coordinates the training process and ensures all components are synchronized; (V) calculates the global model by adding up the noisy local models of the clients and the negative of the aggregated noise from the compensator (unmasking the global model).

The server is in charge of authenticating the coordinator, participants, clients and compensator. The coordinator interacts with the WebApp GUI and is authenticated by the password mechanism. The participants interact with the client GUI and are authenticated via the combination of username, password, project ID, and token before joining the project. The client communicates with the server to send/receive the noisy local/global model parameters, and is authenticated using the combination of username, project ID, and token. To authenticate the compensator, each client shares the hash (\textit{SHA256} algorithm) of the project ID as well as the hash of its username and token with the compensator. The hash of the project ID is employed by the compensator to identify the clients involved in the project. The compensator computes two additional hash values: \textit{username hash} and \textit{token hash}. To calculate the former, it sorts the hash values of the usernames, concatenates them, and takes the hash over the concatenated value. The latter is computed in a similar manner. The server authenticates the compensator via the hash of the project ID, \textit{username hash}, and \textit{token hash} shared with the server.

The server is also responsible for coordinating the training process and synchronization of the clients and compensator. For coordination, the server specifies the \textit{project step} to direct the clients to compute the local parameters of interest in each communication round. For synchronization, the clients share the project step and communication round with the server and compensator. The compensator relays the synchronization parameters to the server, which in turn, makes sure that they are the same for all clients, the compensator, and the server.

\textit{HyFed} can be configured in two operation modes: simulation or federated. In the simulation mode, all components are running in the same workstation and their communication is performed through the loopback interface. In the federated mode (Figure \ref{fig:hyfed-arch}), the WebApp and server are installed in the same workstation while the compensator and client are running in their own machines. The WebApp and server components communicate via the local network. The client-server, client-compensator, and compensator-server communication is carried out over a secure channel using the HTTPS protocol. There is no communication between different machines running the client component.

\begin{figure*}[!ht]
    \centering
    \begin{minipage}{1.0\textwidth}
        \centering
        \includegraphics[width=1\textwidth]{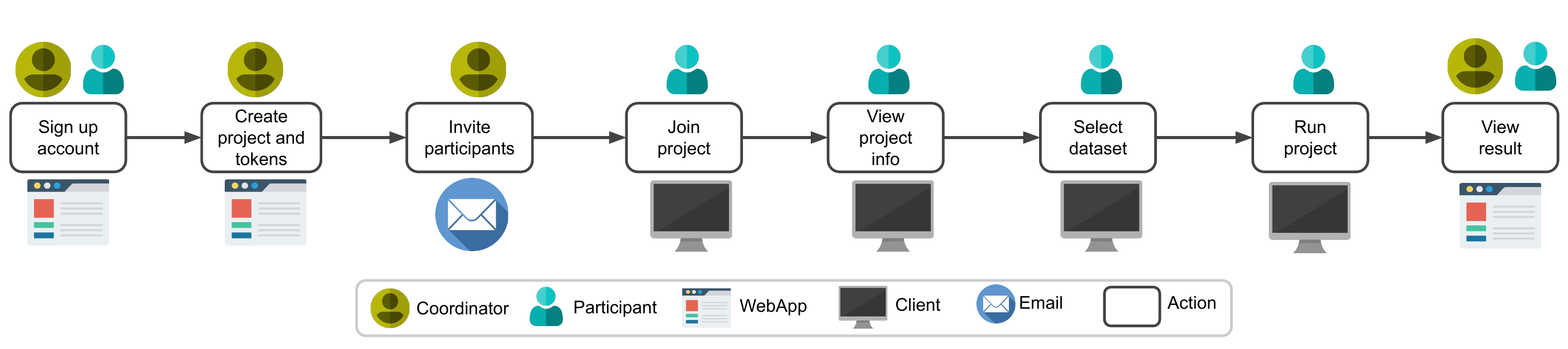}
    \end{minipage}
     \caption{Functional workflow of \textit{HyFed}: The coordinator interacts with the WebApp to create project/tokens while the participants employ the client GUI to join the project, view the project information, select the dataset, and run the project. Both coordinator and participants leverage the WebApp to sign up an account and view the results.
     }
    \label{fig:hyfed-workflow}
\end{figure*}
\subsection{Functional Workflow}
The \textit{functional} workflow of the \textit{HyFed} framework (Figure \ref{fig:hyfed-workflow}) describes the interactions between the human entities (coordinator or participant) and the components (WebApp or client) of the framework step by step and is as follows: (1) The coordinator and each participant sign up for a user account through the WebApp GUI; (2) the coordinator logs into her/his account using the corresponding credentials (i.e. username and password) and creates a new project by specifying the title and description of the project as well as the algorithm and associated hyper-parameters; (3) the coordinator generates a unique token for each participant and shares the project ID (automatically generated) and the token with the participant in a secure manner such as email; (4) the participants employ the client GUI to join the project using the corresponding credentials, project ID, and token; (5) the participants are provided with the project information (e.g. coordinator username, algorithm, and parameters) after joining the project and decide whether or not to proceed; (6) If the participants agree to move forward, they select the dataset file(s) and run the project; (7) the coordinator and participants wait for the project to be completed to view the final results.
\label{subsec:hyfed-workflow}

 \begin{figure*}[!htb]
    \centering
    \begin{minipage}{0.98\textwidth}
\begin{mintedbox}{python}
class VarianceClientProject(HyFedClientProject):
    def compute_local_parameters(self): 
        super().pre_compute_local_parameters() 
        
        if self.project_step == HyFedProjectStep.INIT:
            self.open_data_file()  # Read data file and initialize self.data 
            
        elif self.project_step == "Sum":
            sample_count = self.data.shape[0]
            sample_sum = numpy.sum(self.data, axis=0)
            self.local_parameters["local-count"] = sample_count
            self.local_parameters["local-sum"] = sample_sum
            self.set_compensator_flag({"local-count": DataType.NON_NEGATIVE_INTEGER, "local-sum": DataType.NUMPY_ARRAY_FLOAT})
            
        elif self.project_step == "Sum-square-error":
            global_mean = self.global_parameters["global-mean"]
            sse = numpy.sum(numpy.square(self.data - global_mean), axis=0)
            self.local_parameters["local-sse"] = sse
            self.set_compensator_flag({"local-sse": DataType.NUMPY_ARRAY_FLOAT})
            
        elif self.project_step == HyFedProjectStep.RESULT:
            super().result_step()  
        elif self.project_step == HyFedProjectStep.FINISHED:
            super().finished_step()
            
        super().post_compute_local_parameters()
\end{mintedbox}
    \hfill
    \end{minipage} 

    \begin{minipage}{0.98\textwidth}
\begin{mintedbox}{python}
class VarianceServertProject(HyFedServerProject):
    def aggregate(self):
        super().pre_aggregate() 
        
        if self.step == HyFedProjectStep.INIT:
            self.set_step("Sum")
            
        elif self.step == "Sum":
            self.global_count = self.compute_aggregated_parameter("local-count", DataType.NON_NEGATIVE_INTEGER) 
            global_sum = self.compute_aggregated_parameter("local-sum", DataType.NUMPY_ARRAY_FLOAT) 
            global_mean = global_sum / self.global_count
            self.global_parameters["global-mean"] = global_mean
            self.set_step("Sum-square-error")
            
        elif self.project_step == "Sum-square-error":
            global_sse = self.compute_aggregated_parameter("local-sse", DataType.NUMPY_ARRAY_FLOAT) 
            self.global_variance = global_sse / self.global_count
            self.prepare_results()
            self.set_step(HyFedProjectStep.RESULT)
            
        elif self.project_step == HyFedProjectStep.RESULT:
            super().result_step()  

        super().post_aggregate()
\end{mintedbox}
    \end{minipage}
    \caption{The federated variance algorithm using the \textit{HyFed} client (top) and server (bottom) API: The \textit{import} statements are not shown for simplicity; The assumption is that data file contains multiple columns of numbers, and self.data is initialized with the values of the columns as a \textit{NumPy} array.}
    \label{fig:hyfed-variance}
\end{figure*}

\subsection{Masking Mechanism}
\label{subsec:masking-mech}
\textit{HyFed} can be considered as a FL-SMPC framework, which combines federated learning with \textit{additive secret sharing}-based SMPC. The server and compensator components are the computing parties, the local model is the \textit{secret} of the client, and the client generates two secret shares (i.e. noise and noisy local model) from its secret and shares one of them with compensator and the other with server. The difference is that the server and compensator do not exchange the intermediate results with each other and only the compensator sends them to the server. Given that, the masking mechanism of \textit{HyFed} follows the additive secret sharing rules to hide the original values of the local parameters from a third party.

The local model $M_{L}$ (the original value of a local parameter) is a \textit{non-negative integer}, \textit{negative integer}, or \textit{floating-point number}. \textit{HyFed} employs different masking mechanisms according to $M_{L}$ data type: For non-negative integers, \textit{HyFed} uses the standard additive secret sharing based on finite field arithmetic~\cite{cramer2015smpc}, whereas it leverages \textit{real number secret sharing} based on \textit{Gaussian} distribution~\cite{dibert2014infinite-secret-sharing, tjell2021real-secret-sharing} for floating-point numbers and negative integers.

For non-negative integers, \textit{HyFed} considers a finite field $\mathbb{Z}_{p}$=$\{0,\ 1, \ p-1 \}$, where \textit{p} is a \textit{prime} number. The assumption is that both local model $M_{L}$ and noise $N_{L}$ are integers from the field, i.e. $M_{L}, N_{L} \in \mathbb{Z}_{p}$, and the \textit{addition} and \textit{subtraction} operations are performed using \textit{modular arithmetic} over $\mathbb{Z}_{p}$. To mask local model $M_{L_{j}}$, each client $j$ first generates noise $N_{L_{j}}$, a random integer with uniform distribution from $\mathbb{Z}_{p}$, then computes the noisy local model using modular addition: $M^{\prime}_{L_{j}}$ = $(M_{L_{j}} + N_{L_{j}})$ $\mod$ $p$, and finally shares the noisy local model $M^{\prime}_{L_{j}}$ with the server and noise $N_{L_{j}}$ with the compensator. 

The compensator adds up the noise values from the clients using modular addition to compute the aggregated noise: $N$ = ($\sum_{j=1}^{K} N_{L_{j}}$) $\mod$ $p$. The server first computes the aggregated noisy model by taking sum over the noisy local models of the clients using modular addition, and then, subtracts the aggregated noise from the aggregated noisy model using modular subtraction to calculate (unmask) the global model: $M_{G}$ = ((($\sum_{j=1}^{K} M^{\prime}_{L_{j}}$) $\mod$ $p$) - $N$) $\mod$ $p$.
    
For floating-point numbers and negative integers, each client $j$ generates random noise $N_{L_{j}}$ from the \textit{Gaussian} distribution with \textit{mean of zero} and \textit{variance} of $\sigma_{N}^{2}$, computes the noisy local model: $M^{\prime}_{L_{j}}$ = $M_{L_{j}} + N_{L_{j}}$, and shares $N_{L_{j}}$ with the compensator and $M^{\prime}_{L_{j}}$ with the server. The compensator adds up the noise values of the clients to calculate the aggregated noise $N$ = $\sum_{j=1}^{K} N_{L_{j}}$. The server unmasks the global model by taking sum over the noisy local models and the negative of the aggregated noise:  $M_{G}$ = $\sum_{j=1}^{K} M^{\prime}_{L_{j}} - N$.

The value of prime number $p$ can affect the \textit{correctness} of the masking mechanism. The global models computed with and without masking are the same provided that: (I) $\sum_{j=1}^{K} M_{L_{j}} < p$, and (II) $\sum_{j=1}^{K} N_{L_{j}}$ and $\sum_{j=1}^{K} M^{\prime}_{L_{j}}$ do not cause overflow using 64-bit integers. \textit{HyFed} uses the \textit{NumPy} library~\cite{harris2020numpy} for ordinary and modular addition and subtraction. The largest integer supported by \textit{NumPy} is $64$-bit independent of the platform. A suitable value for $p$ can be determined based on the maximum number of clients. We can set $p$ to the largest prime number that can fit in  ($63$ - $\lceil log_{2}^{K} \rceil$) bits to ensure that the second condition always holds even if all $K$ clients generate the maximum noise (i.e. $p$ - $1$). Notice that the $\mod$ operation in \textit{NumPy} should be used with \textit{signed integer variables}, and as a result, we store non-negative integer values in signed integer variables instead of \textit{unsigned integer variables}, which implies that the largest possible value of $p$ should fit in $63$ bits. The current default value of $p$ is ($2^{54}$ - $33$), which is the largest prime fitting in $54$ bits, and can support maximum of around $500$ clients.  

 Similarly, the value of \textit{variance} $\sigma_{N}^{2}$ might impact the precision of the unmasked global model. This is because HyFed uses $64$-$bit$ float data type of \textit{NumPy} with finite precision for numerical calculations involving floating-point numbers and negative integers, and consequently, too large values of $\sigma_{N}^{2}$ can reduce the precision of the global model. The default value of $\sigma_{N}^{2}$ is $10^{12}$. We discuss the effect of $p$ and $\sigma_{N}^{2}$ on the privacy of the masking mechanism in subsection \ref{subsec:hyfed-privacy}.
 
\subsection{API}
The client, server, and compensator components of \textit{HyFed} have been written in \textit{Python3}. The WebApp has been implemented using \textit{Angular} and \textit{HTML}/\textit{CSS}. The Python-based components employ \textit{NumPy} for computation and aggregation and the \textit{Requests} package to send the parameters over the network. The \textit{Client} \textit{GUI} is based on the \textit{tkinter} package, and the server as well as the compensator leverage the \textit{Django} framework~\cite{django} to implement the back-end functions. The client component leverages \textit{NumPy} to generate the noise values, which are pseudo-random numbers. Notice that a pseudo-random number generator (PRNG) is not cryptographically secure and we chose it for performance reasons. There are cryptographically secure PRNGs, which can generate truly random numbers but with higher overhead in general.

Given the implementation, \textit{HyFed} provides the developers with client-side and server-side Python API to implement their federated algorithms. To this end, the developers: (I) specify the steps and parameters of the algorithm as well as the local and global parameter names of each step, and (II) implement the client-side and server-side logic to compute the local and global parameters corresponding to each step, (III) employ the \textit{HyFed} client API to obtain the global model from the server, mask the local model with noise, and share the noisy local model and noise with the server and compensator, respectively, and (IV) leverage the \textit{HyFed} server API for aggregation and updating the global model. Notice that the compensator component is algorithm-agnostic, and as a result, there is no need for the compensator API. The clients should only know the compensator URL (Uniform Resource Locator) for communication. \textit{HyFed} does not provide the WebApp API and developers should add their algorithm-specific code to the existing WebApp. However, it does not require considerable effort because the WebApp-side code only implements the logic to specify the parameter values of the algorithm, and the associated code size is very small. 

As a simple case study, we outline the development of the \textit{federated variance} algorithm using the \textit{HyFed} API (Figure \ref{fig:hyfed-variance}). The algorithm has no hyper-parameter and consists of three general steps provided by the API (i.e. \textit{Init}, \textit{Result}, and \textit{Finished}), and two algorithm-specific steps: 
\begin{enumerate}
    \item \textit{Init}: The clients open the dataset file and initialize their local data variable.
    \item \textit{Sum}: The clients compute \textit{local-sum} (the sum of the sample values) and \textit{local-count} (number of samples) as the local model parameters; the server performs the aggregation to calculate \textit{global-mean} as the global parameter.
    \item \textit{Sum-square-error}: The clients calculate \textit{local-sse} (the sum of the square error between the sample values and the global mean obtained from the server) as the local parameter; the server computes the \textit{global variance} as the final result.
    \item \textit{Result}: The clients download the result file from the server.
    \item \textit{Finished}: The last step indicating that the project is completed.
\end{enumerate}

To develop the algorithm using the \textit{HyFed} API, we define the \textit{VarianceClientProject} and \textit{VarianceServerProject} classes by inheriting them from the \textit{HyFedClientProject} and \textit{HyFedServerProject} classes, respectively. To implement the client-side/server-side logic, we override the \textit{compute\_local\_parameters}/\textit{aggregate} function in the former/latter. To compute the local parameters on the client side, we can access the global parameters from the server through the \textit{global\_parameters} attribute.  Similarly, we can use the \textit{compute\_aggregated\_parameter} function from the \textit{HyFed} API, which takes the name and data type of the parameter as input, to perform the aggregation on the server side. After computing the values of the local/global parameters, we insert them into \textit{local\_parameters}/\textit{global\_parameters} attributes to share them with the server/client. To hide the local model from the server in the \textit{Sum} and \textit{Sum-square-error} steps, we call the \textit{set\_compensator\_flag} function of the client API in the associated steps.


\label{subsec:hyfed-api}

\subsection{Privacy}
\label{subsec:hyfed-privacy}
The \textit{HyFed} framework assumes \textit{honest-but-curious} clients, server, and compensator, where (I) they adhere to the protocol and completely follow it, and (II) they do not collude with each other, but (III) they leverage any data available to them including the model parameters to reconstruct the private data. By collusion we mean the compensator only shares the aggregated noise but not the individual noise values from the clients with the server; the server never sends the noisy models of the clients to the compensator; the clients send the noise values only to the compensator but not the server and the other clients; similarly, the clients share the noisy local models only with the server but not the other components. Moreover, (IV) there are at least three clients and the clients, server, and compensator components are running in separate physical machines. 

Under the above assumptions,  we discuss the privacy of the \textit{HyFed}'s masking mechanism using the information theoretic measure known as \textit{mutual information} between two random variables $X$and $Y$~\cite{cover1999mutual-information}:
\begin{equation*}
    I(X, Y) = H(X) - H(X|Y)
\end{equation*}
where $H(X)$ is the entropy of $X$ and $H(X|Y)$ is the conditional entropy of $X$ given $Y$. Mutual information indicates the reduction in uncertainty about $X$ given the knowledge of $Y$ and is measured in terms of bits. In our case, $X$ and $Y$ are random variables associated with the distributions of local models ($M_{L}$) and noisy local models ($M^{\prime}_{L}$) of a particular client, respectively. For non-negative integers, it has been shown that the knowledge of $Y$ gives no information about $X$, implying $I(X, Y)$ = $0$~\cite{cramer2015smpc}. In other words, the distributions of local models and noisy local models are indistinguishable, and therefore, sharing the latter does not leak information about the former. Notice that this conclusion holds for any value of $p$.  

For floating-point numbers and negative integers, we assume that the variance of the local models of a particular client is $\sigma^{2}_{M_{L}}$. The distribution of noise values is \textit{Gaussian} with variance of $\sigma^{2}_{N}$. If the the distribution of noisy local models is \textit{Gaussian}  (variance $\sigma^{2}_{M_{L}} + \sigma^{2}_{N}$), then the mutual information between $X$ and $Y$ is maximum~\cite{tjell2021real-secret-sharing}. Therefore, the upper bound on the mutual information is:
\begin{equation*}
    I(X, Y) = \frac{1}{2}\log_{2}(1 + \frac{\sigma^{2}_{M_{L}}}{\sigma^{2}_{N}})
\end{equation*}

That is, the maximum reduction in uncertainty about the local models is determined by the variance of the local models and noise values ($\frac{\sigma^{2}_{M_{L}}}{\sigma^{2}_{N}}$), where the larger values of $\sigma^{2}_{N}$ yield to lower mutual information. Notice that the mean of the \textit{Gaussian} noise generator does not considerably affect the privacy, and therefore, we set it to zero according to \cite{tjell2021real-secret-sharing}. Moreover, the default value of $\sigma^{2}_{N} = 10 ^{12}$ is a very large variance, which can provide negligible mutual information for most algorithms.

It is worth mentioning that \textit{HyFed} does not offer output privacy because the global parameters are finally released, which might be employed to extract the private information of the individuals. Using differential privacy is a plausible solution to this issue but at the expense of the global model utility.

\section{Conclusion}
\label{sec:conclusion}
In this paper, we introduced \textit{HyFed} as a generic and open-source framework, comprising the WebApp, client, compensator, and server software components, which can run in the simulation or federated mode. Using the federated variance algorithm as a case study, we demonstrated how developers can implement their federated and privacy-enhanced algorithms via the \textit{HyFed} generic API. The \textit{HyFed} framework assumes honest-but-curious threat model and improves the privacy of FL while preserving the utility of the global model.

As a future direction, we plan to (I) extend the \textit{HyFed} framework and API to allow for integration with popular machine learning and deep learning libraries, (II) deploy the federated tools in the real-world settings and conduct thorough experiments to evaluate the scalability of the framework, (III) provide the WebApp API to avoid modifying the current WebApp code to implement customized algorithms, and (IV) add the monitoring capabilities to the \textit{HyFed} framework to measure the runtime as well as the network bandwidth usage of the algorithms.

\nocite{langley00}

\bibliography{hyfed_references}
\bibliographystyle{sysml2019}


\end{document}